\documentclass[11pt,letterpaper]{article}
\usepackage{naaclhlt2016}
\usepackage{times}
\usepackage{latexsym}

\usepackage{url}
\usepackage{latexsym}
\usepackage{multirow}
\usepackage{rotating}
\usepackage{amsmath}
\usepackage{amsthm}
\usepackage{amssymb}
\usepackage{colortbl}
\usepackage{booktabs}
\usepackage{tabularx}
\usepackage{paralist}
\usepackage{graphicx}
\usepackage{mathtools}
\usepackage{footmisc}
\usepackage{lpic}
\usepackage[utf8]{inputenc}
\usepackage{titlesec}

\naaclfinalcopy 
\def\naaclpaperid{38} 

\title{Ultradense Word Embeddings by Orthogonal Transformation}

\author{Sascha Rothe \and Sebastian Ebert \and Hinrich Schütze\\
Center for Information and Language Processing\\
LMU Munich, Germany\\
\url{{sascha|ebert}@cis.lmu.de}}

\date{}

\def\figref#1{Figure~\ref{fig:#1}}
\def\figlabel#1{\label{fig:#1}\label{p:#1}}

\def\tabref#1{Table~\ref{tab:#1}}
\def\tablabel#1{\label{tab:#1}\label{p:#1}}

\def\secref#1{Section~\ref{sec:#1}}
\def\seclabel#1{\label{sec:#1}\label{p:#1}}
\def\eqref#1{Eq.~\ref{eqn:#1}}

\def\eqsref#1#2{Eqs.~\ref{eqn:#1}/\ref{eqn:#2}}
\def\eqlabel#1{\label{eqn:#1}}

\long\def\symbolfootnote[#1]#2{\begingroup%
\def\thefootnote{\fnsymbol{footnote}}\footnote[#1]{#2}\endgroup}

\newcommand{\sigd}{${}^\dagger$}
\newcommand{\signo}{\textcolor{white}{${}^\dagger$}}

\newcounter{notecounter}
\newcommand{\enotesoff}{\long\gdef\enote##1##2{}}
\newcommand{\enoteson}{\long\gdef\enote##1##2{{
\stepcounter{notecounter}
\large\bf
\hspace{1cm}\arabic{notecounter} $<<<$ ##1: ##2
$>>>$\hspace{1cm}}}}
\enoteson
\enotesoff

\newcounter{notepriotwocounter}
\newcommand{\enotepriotwosoff}{\long\gdef\enotepriotwo##1##2{}}
\newcommand{\enotepriotwoson}{\long\gdef\enotepriotwo##1##2{{
\stepcounter{notepriotwocounter}
\large\bf
\hspace{1cm}\arabic{notepriotwocounter} $<<<$ ##1: ##2
$>>>$\hspace{1cm}}}}
\enotepriotwoson
\enotepriotwosoff

\newcommand{\densifier}{\textsc{Densifier}}

\begin{document}

\maketitle

\begin{abstract}
Embeddings are \emph{generic} representations that are useful for many NLP tasks.  In this paper, we introduce \densifier{}, a method that learns an \emph{orthogonal transformation} of the embedding space that focuses the information relevant for a task in an \emph{ultradense subspace} of a dimensionality that is smaller by a factor of 100 than the original space. We show that ultradense embeddings generated by \densifier{} reach state of the art on a lexicon creation task in which words are annotated with three types of lexical information -- sentiment, concreteness and frequency. On the SemEval2015 10B sentiment analysis task we show that no information is lost when the ultradense subspace is used, but training is an order of magnitude \emph{more efficient} due to the compactness of the ultradense space.
\end{abstract}

\section{Introduction}\seclabel{intro}
Embeddings are useful for many tasks, including word similarity (e.g., \newcite{PenSocMan14}), named entity recognition (NER) (e.g., \newcite{collobert11scratch}) and sentiment analysis (e.g., \newcite{Kim14}, \newcite{KalGreBlu14}, \newcite{SevMos15}). Embeddings are generic representations, containing different types of information about a word. Statistical models can be trained to make best use of these generic representations for a specific application like NER or sentiment analysis \cite{EbeVuSch15b}.

Our hypothesis in this paper is that the information useful for any given task is contained in an \emph{ultradense subspace} $E_u$. We propose the new method \densifier{} to identify $E_u$. Given a set of word embeddings, \densifier{} learns an \emph{orthogonal transformation} of the original space $E_o$ on a task-specific training set.
The orthogonality of the transformation can be considered a
hard regularizer. 

The benefit of the proposed method is that embeddings are most useful if learned on unlabeled corpora and performance-enhanced on a broad array of tasks. This means we should try to keep all information offered by them.
Orthogonal transformations ``reorder'' the space without
adding or removing information and preserve the bilinear
form, i.e., Euclidean distance  and cosine. The transformed
embeddings concentrate all information relevant for the task
in $E_u$. 

The benefits of $E_u$ compared to $E_o$ are (i) \emph{high-quality} and (ii) \emph{efficient} representations.
(i) \densifier{} moves non-task-related information outside of $E_u$, i.e., into the orthogonal complement of $E_u$. As a result,  $E_u$ provides \emph{higher-quality representations} for the task than $E_o$; e.g., noise that could result in overfitting is reduced in  $E_u$ compared to $E_o$.
(ii) $E_u$ has a \emph{dimensionality smaller by a factor of 100} in our experiments. As a result, training statistical models on these embeddings is much faster. These models also have many fewer parameters, thus again helping to prevent overfitting, especially for complex, deep neural networks.

We show the benefits of ultradense representations in two text polarity classification tasks (SemEval2015 Task 10B, Czech movie reviews).

In the most extreme form, ultradense representations -- i.e., $E_u$ -- have a single dimension. We exploit this for creating lexicons in which words are annotated with lexical information, e.g., with sentiment. Specifically, we create high-coverage lexicons with up to 3 million words
(i) for three lexical properties: for \emph{sentiment}, \emph{concreteness} and \emph{frequency};
(ii) for five languages: \emph{Czech}, \emph{English}, \emph{French}, \emph{German} and \emph{Spanish};
(iii) for two domains, \emph{Twitter} and \emph{News}, in a domain adaptation setup.

The main advantages of this method of lexicon creation are:
(i) We need a training lexicon of only a few hundred words, thus making our method effective for new domains and languages and requiring only a minimal manual annotation effort.
(ii) The method is applicable to any set of embeddings, including phrase and sentence embeddings. Assuming the availability of a small hand-labeled lexicon, \densifier{} automatically creates a domain dependent lexicon based on a set of embeddings learned on a large corpus of the domain.
(iii) While the input lexicon is \emph{discrete}  -- e.g., positive (+1) and negative (-1) sentiment -- the output lexicon is \emph{continuous} and this more fine-grained assessment is potentially more informative than a simple binary distinction.

We show that lexicons created by \densifier{} beat the state of the art on SemEval2015 Task 10E (determining association strength).

\enotepriotwo{hs}{last benefit: do we show anywhere that this works?}\enotepriotwo{sr}{the task determining association strength is build on this fact}

One of our goals is to make embeddings more interpretable. The work on sentiment, concreteness and frequency we present in this paper is a first step towards a \emph{general decomposition of embedding spaces into meaningful, dense subspaces}. This would lead to cleaner and more easily interpretable representations -- as well as representations that are more effective and efficient.

\section{Model}\seclabel{model}
Let $Q \in \mathbb{R}^{d \times d}$ be an orthogonal matrix
that transforms the original word embedding space into a
space in which certain types of information are represented
by a small number of dimensions. Concretely, we learn $Q$
such that the dimensions $D^s \subset \{1, \ldots, d\}$ of
the resulting space correspond to a word's sentiment
information and the $\{1, \ldots, d\} \setminus D^s$
remaining dimensions correspond to non-sentiment
information. Analogously, the sets of dimensions $D^c$ and
$D^f$ correspond to a word's concreteness information and
frequency information, respectively. In this paper, we
assume that these properties do not correlate and therefore
the ultradense subspaces do not overlap, e.g., $D^s \cap D^c
= \emptyset$. However, this might not be true for other
settings, e.g., sentiment and semantic information.

If $e_w \in E_o \subset \mathbb{R}^{d}$ is the original embedding of word $w$, the transformed representation is $Q e_w$. We use $*$ as a placeholder for $s$, $c$ and $f$ and call $d^{*} = |D^{*}|$  the dimensionality of the ultradense subspace of $*$. For each ultradense subspace, we create $P^{*} \in \mathbb{R}^{d^{*} \times d}$, an identity matrix for the dimensions in $D^{*} \subset \{1, \ldots, d\}$. Thus, the ultradense representation $u^{*}_w \in E_u \subset \mathbb{R}^{d^{*}}$ of $e_w$ is defined as: 
\begin{equation}
u^{*}_w \coloneqq P^{*} Q e_w
\end{equation}

\subsection{Separating Words of Different Groups}
We assume to have a lexicon resource $l$ in which each word $w$ is annotated for a certain information as either $l^{*}(w)=+1$ (positive, concrete, frequent) or $l^{*}(w)=-1$ (negative, abstract, infrequent). Let $L^{*}_{\not\sim}$ be a set of word index pairs $(v,w)$ for which $l^{*}(v)\neq l^{*}(w)$ holds. We want to maximize:
\begin{equation}
\eqlabel{objective_max} \sum_{(v,w) \in L^{*}_{\not\sim}} \|u^{*}_v - u^{*}_w\|
\end{equation}
Thus, our objective is given by:
\begin{equation}
\eqlabel{objective_max_max} \underset{Q}{\operatorname{argmax}} \sum_{(v,w) \in L^{*}_{\not\sim}} \|P^{*} Q (e_w - e_v)\|
\end{equation}
or, equivalently, by:
\begin{equation}
\mbox{\hspace{1.0cm}}\underset{Q}{\operatorname{argmin}} \sum_{(v,w) \in L^{*}_{\not\sim}} -\|P^{*} Q (e_w - e_v)\|
\end{equation}
subject to $Q$ being an orthogonal matrix.

\subsection{Aligning Words of the Same Group}
Another goal is to minimize the distance of two words of the same group. Let $L^{*}_{\sim}$ be a set of word index pairs $(v,w)$ for which $l^{*}(v)=l^{*}(w)$ holds. In contrast to \eqref{objective_max_max}, we now want to minimize each distance. Thus, the objective is given by:
\begin{equation}
\eqlabel{objective_min} \underset{Q}{\operatorname{argmin}} \sum_{(v,w) \in L^{*}_{\sim}} \|P^{*} Q (e_w - e_v)\|
\end{equation} subject to $Q$ being an orthogonal matrix.

\begin{figure*}[tbhp]
  \centering
  \includegraphics[width=\textwidth]{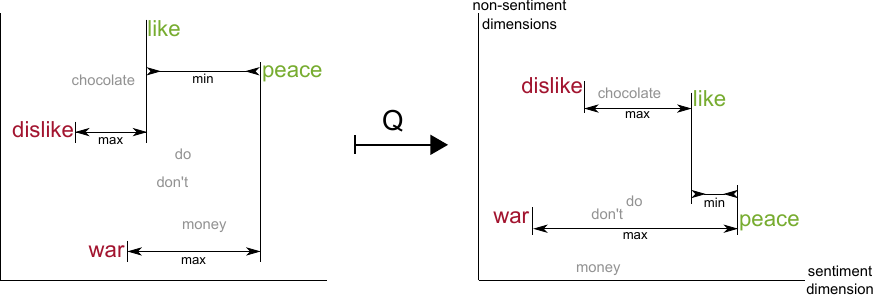}
  \caption{The original word embedding space (left) and the transformed embedding space (right). The training objective for $Q$ is to \emph{minimize} the distances in the sentiment dimension between words of the same group (e.g., positive/green: ``like'' \& ``peace'') and to \emph{maximize} the distances between words of different groups (e.g., negative/red \& positive/green: ``war'' \& ``peace''; not necessarily antonyms).}
  \figlabel{orthogonal_transformation}
\end{figure*}

The intuition behind the two objectives is graphically depicted in \figref{orthogonal_transformation}.

\subsection{Training}
We combine the two objectives in \eqsref{objective_max_max}{objective_min} for each subspace, i.e., for sentiment, concreteness and frequency, and weight them with $\alpha^{*}$ and $1-\alpha^{*}$. Hence, there is one hyperparameter $\alpha^{*}$ for each subspace. We then perform stochastic gradient descent (SGD). Batch size is $100$ and starting learning rate is $5$,  multiplied by $.99$ in each iteration.

\subsection{Orthogonalization}
Each step of SGD updates $Q$. The updated matrix $Q'$ is in
general no longer orthogonal. We therefore reorthogonalize
$Q'$ in each step based on singular value decomposition: \[Q'
= U S V^T\] where $S$ is a diagonal matrix, and $U$ and $V$
are orthogonal matrices. The matrix \[Q \coloneqq U V^T\] is
the nearest orthogonal matrix to $Q'$ in both the 2-norm and
the Frobenius norm~\cite{fan1955some}. (Formalizing
  our regularization directly as projected gradient descent
  would be desirable. However, gradient descent includes an
  additive operation and orthogonal matrices are not closed
  under summation.)

SGD for this problem is sensitive to the learning rate. If the learning rate is too large, a large jump results and the reorthogonalized matrix $Q$ basically is a random new point in the parameter space. If the learning rate is too small, then learning can take long. We found that our training regime of starting at a high learning rate (5) and multiplying by .99 in each iteration is effective. Typically, the cost initially stays about constant (random jumps in parameter space), then cost steeply declines in a small number of about 50 iterations (sweet spot); the curve  flattens after that. Training $Q$ took less than 5 minutes per experiment for all experiments in this paper.

\section{Lexicon Creation}\seclabel{lexicon_creation}
For lexicon creation, the input is a set of \emph{embeddings} and a lexicon \emph{resource} $l$, in which words are annotated for a lexical information such as sentiment, concreteness or frequency.  \densifier{} is then trained to produce a \emph{one-dimensional ultradense subspace}. The output is an \emph{output lexicon}. It consists of all words covered by the embedding set, each associated with its one-dimensional ultradense subspace representation (which is simply a real number), an indicator of the word's strength for that information.

\begin{table*}[tbh]
\small
\center
\begin{tabular}{r|l@{\hspace{0.1cm}}l@{\hspace{0.1cm}}l@{\hspace{0.1cm}}r@{\hspace{0.1cm}}r|l@{\hspace{0.1cm}}r@{\hspace{0.1cm}}r|lrr|r}
 &&  &  & &  & \multicolumn{3}{c|}{train} & \multicolumn{3}{c|}{test} &  \\
&&&&\#tokens&\#types&resource&\multicolumn{1}{c}{$\cap$}&\multicolumn{1}{c|}{\#words}&resource&\multicolumn{1}{c}{$\cap$}&\multicolumn{1}{c|}{\#words}&$\tau$\\\hline
1 & sent & CZ & web & 2.44 & 3.3 & SubLex 1.0 &\footnotesize{2,492}&\footnotesize{4,125} & SubLex 1.0 &\footnotesize{319}&\footnotesize{500} & .580  \\                     
2 & sent & DE & web & 1.34 & 8.0 & German PC &\footnotesize{10,718}&\footnotesize{37,901} & German PC &\footnotesize{573}&\footnotesize{1,000} & .654  \\                   
3 & sent & ES & web & 0.37 & 3.7 & full-strength &\footnotesize{824}&\footnotesize{1,147} & full-strength &\footnotesize{185}&\footnotesize{200} & .563  \\           
4 & sent & FR & web & 0.12 & 1.6 & FEEL &\footnotesize{7,496}&\footnotesize{10,979} & FEEL &\footnotesize{715}&\footnotesize{1,000} & .544  \\                            
5 & sent & EN & twitter & 3.34 & 5.4 & WHM all &\footnotesize{12,601}&\footnotesize{19,329} & Trial 10E &\footnotesize{198}&\footnotesize{200} & .661  \\          
6 & sent & EN & news & 3.00 & 100.0 & WHM train &\footnotesize{7,633}&\footnotesize{10,270} & WHM val &\footnotesize{952}&\footnotesize{1,000} & .622  \\                                   
7 & conc & EN & news & 3.00 & 100.0 & BWK &\footnotesize{14,361}&\footnotesize{29,954} & BWK &\footnotesize{8,694}&\footnotesize{10,000} & .623  \\
8 & freq & EN & news & 3.00 & 100.0 & word2vec order &\footnotesize{4,000}&\footnotesize{4,000} & word2vec order &\footnotesize{1,000}&\footnotesize{1,000} & .361  \\
9 & freq & FR & web & 0.12 & 1.6 & word2vec order &\footnotesize{4,000}&\footnotesize{4,000} & word2vec order &\footnotesize{1,000}&\footnotesize{1,000} & .460  
\end{tabular}
\caption{Results for lexicon creation. \#tokens: size of embedding training corpus (in billion). \#types: size of output lexicon (in million). For each resource, we give its size (``\#words'') and the size of the intersection of resource and embedding set (``$\cap$''). Kendall's $\tau$ is computed on ``$\cap$''.}
\tablabel{lexica}
\end{table*}

The embeddings and lexicon resources used in this paper cover five languages and three domains (\tabref{lexica}). The Google News embeddings for English\footnote{\url{https://code.google.com/p/word2vec/}} and the FrWac embeddings for French\footnote{\url{http://fauconnier.github.io/}} are publicly available. We use word2vec to train 400-dimensional embeddings for English on a 2013 Twitter corpus of size 5.4 $\times 10^9$. For Czech, German and Spanish, we train embeddings on web data of sizes 3.3, 8.0 and 3.8 $\times 10^9$, respectively. We use the following lexicon resources for sentiment: SubLex 1.0~\cite{VesBoj13} for Czech; WHM for English [the combination of MPQA~\cite{wilson2005recognizing}, Opinion Lexicon~\cite{HuLiu04} and NRC Emotion lexicons~\cite{MohTur13}]; FEEL~\cite{AbdAzeBri+14} for French; German Polarity Clues~\cite{Wal10} for German; and the sentiment lexicon of \newcite{PerBanMih12} for Spanish. For concreteness, we use BWK, a lexicon of 40,000 English words \cite{brysbaert2014concreteness}. For frequency, we exploit the fact that word2vec stores words in frequency order; thus, the ranking provided by word2vec is our lexicon resource for frequency. 

\enotepriotwo{hs}{what was the ``validation'' set i am now calling
  the ``test set''. was there a particular reason why oyu
  wanted to call it a validation set?}\enotepriotwo{sr}{parameters are tuned on this set}

For a resource/embedding-set pair $(l,E)$, we intersect the vocabulary of $l$ with the top 80,000 words of $E$ to filter out noisy, infrequent words that tend to have low quality embeddings and we do not want them to introduce noise when training the transformation matrix.

For the sentiment and concreteness resources, $l^*(w) \in
\{-1,1\}$ for all words $w$ covered. We create a resource
$l^f$ for frequency by setting $l^f(w) = 1$ for the 2000
most frequent words  and $l^f(w) = -1$ for  words at ranks
20000-22000. 1000 words randomly selected from the 5000 most
frequent are the test set.\footnote{The main result of the
  frequency experiment below is that $\tau$ is low even in a
  setup that is optimistic due to train/test overlap;
  presumably it would be even lower without overlap.} We
designate three sets of dimensions $D^s$, $D^c$ and $D^f$
to represent sentiment, concreteness and frequency,
respectively, and arbitrarily set (i) $D^c=\{11\}$ for
English and $D^c=\emptyset$ for the other languages since we
do not have concreteness resources for them, (ii)
$D^s=\{1\}$ and (iii) $D^f=\{21\}$. Referring to the lines in \tabref{lexica}, we then learn six orthogonal transformation matrices $Q$: for CZ-web (1), DE-web (2), ES-web (3), FR-web (4, 9), EN-twitter (5) and EN-news (6, 7, 8).

\section{Evaluation}\seclabel{evaluation}

\subsection{Top-Ranked Words}
\tabref{sentiment_lexicon} shows the top 10 positive/negative words (i.e., most extreme values on dimension $D^{s}$) when we apply the transformation to the corpora EN-twitter, EN-news and DE-web and the top 10 concrete/abstract words (i.e., most extreme values on dimension $D^c$) for EN-news. For EN-twitter (leftmost double column), the selected words look promising: they contain highly domain-specific words such as hashtags (e.g., \#happy). This is surprising because there is not a single hashtag in the lexicon resource WHM that \densifier{} was trained on. Results for the other three double columns show likewise extreme examples for the corresponding information and language. This initial evaluation indicates that our method effectively learns high quality lexicons for new domains.  \figref{visualization} depicts values for selected words for the three properties. Illustrative examples are ``brother'' / ``brotherhood'' for concreteness and ``hate'' / ``love'' for sentiment.

\begin{table*}[tbhp]
\small
\center
\begin{tabular}{cc|cc|cc|cc}
\multicolumn{2}{c|}{EN-twitter} & \multicolumn{2}{c|}{EN-news} & \multicolumn{2}{c|}{EN-news} & \multicolumn{2}{c}{DE-web} \\
positive & negative & positive & negative & concrete & abstract & positive & negative \\\hline
\#blessed & rape & expertise & angry & tree & fundamental & herzlichen & gesperrt \\
inspiration & racist & delighted & delays & truck & obvious & kenntnisse & droht \\
blessed & horrible & honored & worse & kitchen & legitimate & hervorragende & verurteilt \\
inspiring & nasty & thank & anger & dog & reasonable & ideale & gefahr \\
foundation & jealousy & wonderful & foul & bike & optimistic & bestens & falsche \\
provide & murder & commitment & blamed & bat & satisfied & glückwunsch & streit \\
wishes & waste & affordable & blame & garden & surprising & optimale & angst \\
dedicated & mess & passion & complained & homer & honest & anregungen & krankheit \\
offers & disgusting & exciting & bad & bed & regard & freuen & falschen \\
\#happy & spam & flexibility & deaths & gallon & extraordinary & kompetenzen & verdacht 
\end{tabular}
\caption{Top 10 words in the output lexicons for the domains
  Twitter and
  News (English) and Web (German).}
\tablabel{sentiment_lexicon}
\end{table*}

\def\graphfactor{0.7}

\subsection{Quality of Predictions}
\tabref{lexica} presents experimental results. In each case, we split the resource into train/test, except for Twitter where we used the trial data of SemEval2015 Task 10E for test. We train \densifier{} on train and compute Kendall's $\tau$ on test. The size of the lexicon resource has no big effect; e.g., results for Spanish (small resource; line 3) and French (large resource; line 4) are about the same. See \secref{lexicon_size_analysis} for a more detailed analysis of the effect of resource size.

The quality of the output lexicon depends strongly on the quality of the underlying word embeddings; e.g., results for French (small embedding training corpus; line 4) are worse than results for English  (large embedding training corpus; line 6) even though the lexicon resources have comparable size.
\enotepriotwo{se}{That might also be an effect of the embeddings domain. FR-web vs. EN-news, but both lexicons are non-web.}

In contrast to sentiment/concreteness, $\tau$ values for frequency are low (lines 8-9). For the other three languages we obtain $\tau \in [.34,.46]$ for frequency (not shown). This suggests that word embeddings represent sentiment and concreteness much better than frequency. The reason for this likely is the learning objective of word embeddings: modeling the context. Infrequent words can occur in frequent contexts. Thus, the
frequency information in a single word embedding is limited. In contrast negative words are likely to occur in negative contexts. 

The nine output lexicons in \tabref{lexica} -- each
a list of words annotated with predicted
strength on one of three properties -- are available at
\url{www.cis.lmu.de/~sascha/Ultradense/}.

\begin{figure*}
\centering
  \includegraphics[width=\graphfactor\textwidth]{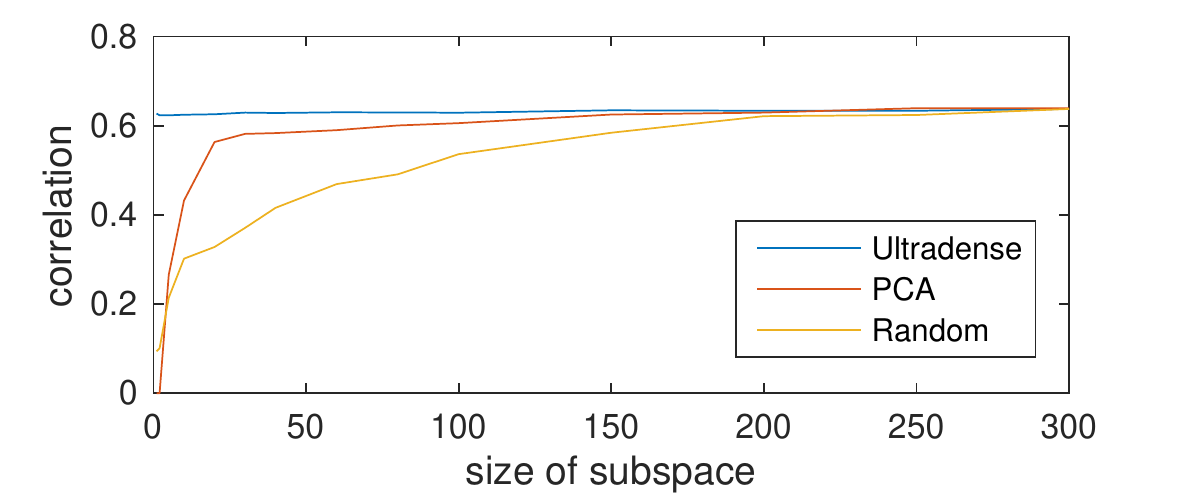}
  \includegraphics[width=\graphfactor\textwidth]{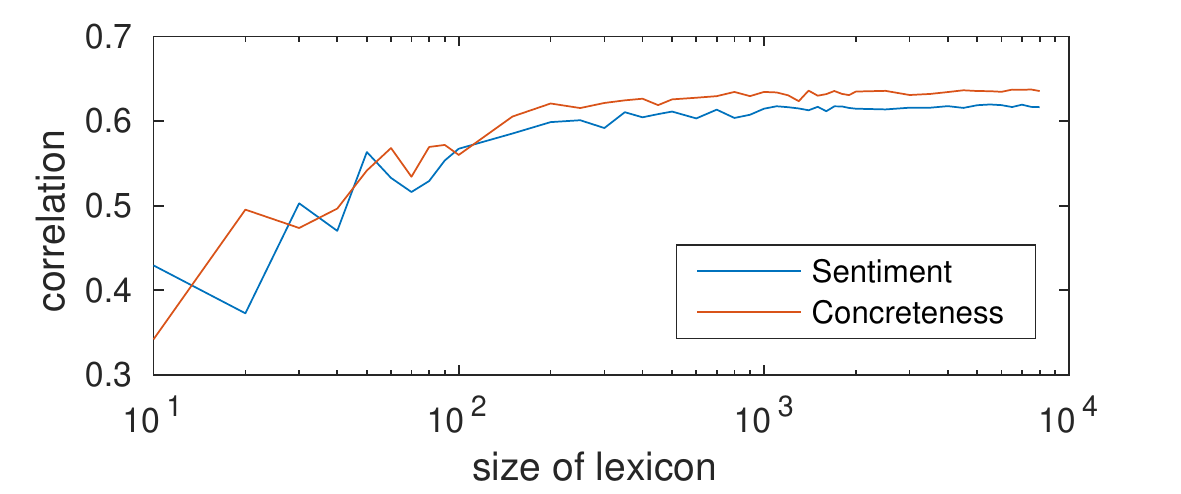}
  \caption{Kendall's $\tau$ versus subspace size (top) and training resource size (bottom). See lines 6 \& 8,  \tabref{lexica}, for train/test split.}
  \figlabel{subspace_dimensions}
  \figlabel{lexicon_size}
\end{figure*}

\subsection{Determining Association Strength}
We also evaluate lexicon creation on SemEval2015 Task 10E. As before, the task is to predict the sentiment score of words/phrases. We use the trial data of the task to tune the hyperparameter, $\alpha^s =$ .4. Out-of-vocabulary words were predicted as neutral (7/1315). \tabref{semeval2015_taskE} shows that the lexicon computed by \densifier{} (line 5, \tabref{lexica}) has a $\tau$ of .654 (line 6, column \emph{all}), significantly better than all other systems, including the winner of SemEval 2015 ($\tau =$ .626, line 1). \densifier{} also beats Sentiment140~\cite{MohKirZhu13}, a widely used semi-automatic  sentiment lexicon. The last column is $\tau$ on the intersection of \densifier{} and Sentiment140. It shows that \densifier{} again performs significantly better than Sentiment140.

\begin{table}
\small
\centering
\begin{tabular}{rlccc}
 &  & & \multicolumn{2}{c}{$\tau$}\\
 & system &  &all & $\cap$ \\\hline\hline
1 & \multicolumn{2}{l}{\newcite{amir-EtAl:2015:SemEval}}		& .626\sigd & \\
2 & \multicolumn{2}{l}{\newcite{hamdan-bellot-bechet:2015:SemEval1}}	& .621\sigd & \\
3 & \multicolumn{2}{l}{\newcite{zhang-wu-lan:2015:SemEval}}		& .591\sigd & \\
4 & \multicolumn{2}{l}{\newcite{ozdemir-bergler:2015:SemEval}}		& .584\sigd & \\
5 & \multicolumn{2}{l}{\newcite{plotnikova-EtAl:2015:SemEval2}}		& .577\sigd & \\\hline
6  & \densifier{}	& & \textbf{.654}\signo & \textbf{.650}\signo\\
7  & Sentiment140 	& & .508\sigd & .538\sigd \\
8  & \densifier{}, trial only && .627\sigd 
\end{tabular}
\caption{Results for Lexicon Creation. The first column
  gives the correlation with the entire test lexicon of
  SemEval2015 10E, the last column only on the intersection
  of our output lexicon and Sentiment140. Of the 1315 words
  of task 10E, 985 and 1308 are covered  by \densifier{} and
  Sentiment140, respectively. $\dagger$: significantly worse
  than the best (bold) result in the same column ($\alpha = .05$, Fisher z-transformation).\tablabel{semeval2015_taskE}}
\end{table}

\subsection{Text Polarity Classification}
We now show that ultradense embeddings decrease model
training times without any noticeable decrease in
performance compared to the original embeddings. We evaluate
on SemEval2015 Task 10B, classification of Twitter tweets as
positive, negative or neutral. We reimplement the
linguistically-informed convolutional neural network
(lingCNN) of \newcite{EbeVuSch15b} that has close to
state-of-the-art performance on the task. We do not use sentence-based features to focus on the evaluation of the embeddings. We initialize the first layer of lingCNN, the embedding layer, in three different ways: (i) 400-dimensional Twitter embeddings (\secref{lexicon_creation}); (ii) 40-dimensional ultradense embeddings derived from (i); (iii) 4-dimensional ultradense embeddings derived from (i). The objective weighting is $\alpha^s=.4$, optimized on the development set.

The embedding layer converts a sentence into a matrix of word embeddings.  We also add linguistic features for words, such as sentiment lexicon scores.  The combination of embeddings and linguistic features is the input for a convolution layer with filters spanning 2-5 words (100 filters each).  This is followed by a max pooling layer,  a rectifier nonlinearity~\cite{NaiHin10} and a fully connected softmax layer predicting the final label.  The model is trained with SGD using AdaGrad~\cite{DucHazSin11} and $\ell_2$ regularization ($\lambda = 5 \times 10^{-5}$). Learning rate is $0.01$. Mini-batch size is 100.

We follow the official guidelines and use the SemEval2013 training and development sets as training set, the SemEval2013 test set as development set and the SemEval2015 test set to report final scores~\cite{NakRosKoz+13,RosNakKir+15}. We report macro $F_1$ of positive and negative classes (the official SemEval evaluation metric) and accuracy over the three classes. \tabref{semeval2015_taskB} shows that 40-dimensional ultradense embeddings perform almost as well as the full 400-dimensional embeddings (no significant difference according to sign test).  Training time is shorter by a factor of 21 (85/4 examples/second). The 4-dimensional ultradense embeddings lead to only a small loss of 1.5\% even though the size of the embeddings is smaller by a factor of 100 (again not a significant drop). Training time is shorter by a factor of 44 (178/4).

We perform the same experiment on CSFD, a Czech movie review dataset \cite{HabPtaSte13}, to show the benefits of ultradense embeddings for a low-resource language where only one rather small lexicon is available. As original word embeddings we train new 400 dimensional embeddings on a large Twitter corpus (3.3 $\times 10^9$ tokens). We use \densifier{} to create 40 and 4 dimensional embeddings out of these embeddings and SubLex 1.0~\cite{VesBoj13}. Word polarity features are also taken from SubLex. A simple binary negation indicator is implemented by searching for all tokens beginning with ``ne''. Since that includes superlative forms having the prefix ``nej'', we remove them with the exception of common negated words, such as ``nejsi'' -- ``you are not''. We randomly split the 91,000 dataset instances into 90\% train and 10\% test  and report accuracy and macro $F_1$ score over all three classes.

\tabref{semeval2015_taskB} shows that what we found for English is also true for Czech. There is only a small performance drop when using ultradense embeddings (not significant for 40 dimensional embeddings) while the speed improvement is substantial.

\begin{table}
\small
\centering
\begin{tabular}{llrrrr}
lang. & embeddings & \#dim & acc & $F_{1}$ & ex./sec\\\hline
en & original & 400 & .666 & .623 & 4 \\
en & \densifier{} & 40 & .662 & .620 & 85 \\
en & \densifier{} &  4 & .646 & .608 & 178 \\\hline
cz & original & 400 & .803 & .802 & 1 \\
cz & \densifier{} & 40 & .803 & .801 & 24 \\
cz & \densifier{} &  4 & .771 & .769 & 83 \\
\end{tabular}
\caption{Performance on Text Polarity Classification}
\tablabel{semeval2015_taskB}
\end{table}

\begin{figure*}[t!]
  \includegraphics[width=\textwidth]{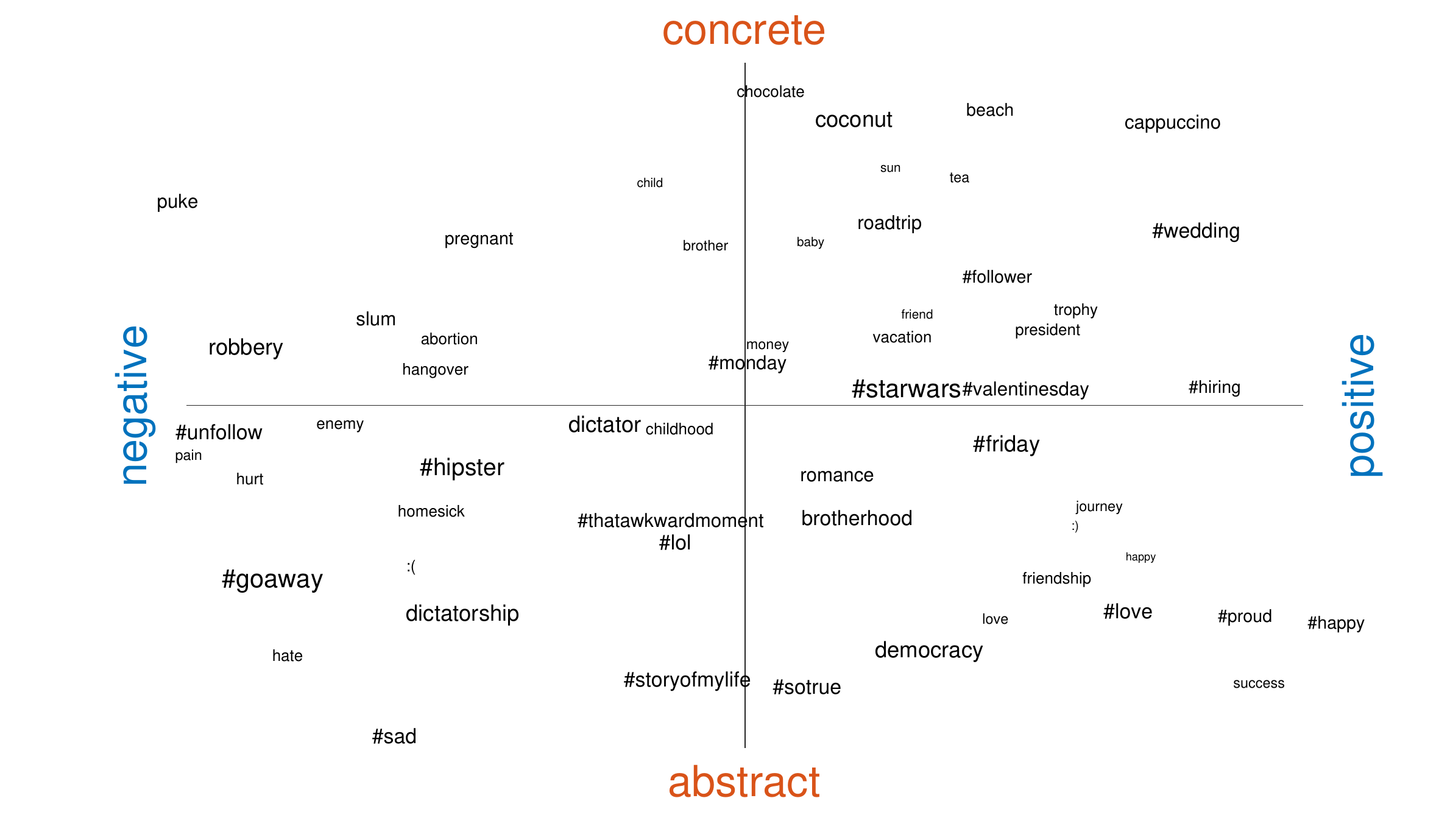}
  \caption{Illustration of EN-twitter output lexicon: \densifier{} values are  x
    coordinate (sentiment), y coordinate (concreteness) and
    font size (frequency)}
  \figlabel{visualization}
\end{figure*}

\section{Parameter Analysis}
In this section, we analyze two parameters: size of ultradense subspace and  size of  lexicon resource. We leave an evaluation of another parameter, the size of the embedding training corpus, for future work, but empirical results suggest that this corpus should ideally have a size of several billion tokens.

\subsection{Size of Subspace}
With the exception of the two text polarity classification experiments, all our subspaces have dimensionality $d^{*}=1$. The question arises: does a one-dimensional space perhaps have too low a capacity to encode all relevant information and could we further improve our results by increasing the dimensionality of the subspace to values $d^{*}>1$? The lexicon resources that we train and test on are all binary; thus, if we use values $d^{*}>1$, then we need to map the subspace embeddings to a one-dimensional scale for evaluation. We do this by training, on the train part of the resource, a linear transformation from the ultradense subspace to the one-dimensional scale (e.g., to the sentiment scale).

\figref{subspace_dimensions} compares different values of $d^s$ for three different types of subspaces in this setup, i.e., the setup in which the subspace representations are mapped via linear transformation to a one-dimensional sentiment value: (i) \emph{Random}: we take  the first $d^s$ dimensions of the original embeddings; (ii) \emph{PCA}: we compute a PCA and take the first $d^s$ principal components; (iii) \emph{Ultradense} subspace of dimensionality $d^s$. We use the word embeddings and lexicon resources of line 6 in \tabref{lexica}. For random, the performance starts dropping when the subspace is smaller than $200$ dimensions. For PCA, the performance is relatively stable until the subspace becomes smaller than $100$. In contrast, ultradense subspaces have almost identical performance for all values of $d^s$, even for $d^s=1$.  This suggests that a single dimension is sufficient to encode all sentiment information needed for sentiment lexicon creation. However, for other sentiment tasks more dimensions may be needed, e.g., for modeling different emotional dimensions of polarity: fear, sadness, anger etc.

An alternative approach to create a low-dimensional space is to simply train low-dimensional word2vec embeddings. The following experiment suggests that this does not work very well.  We used word2vec to train 60-dimensional twitter embeddings with the same settings as on line 5 in \tabref{lexica}. While the correlation for 400-dimensional embeddings shown in \tabref{lexica} is .661, the correlation of 60-dimensional embeddings is only .568. Thus, even though we show that the information in 400-dimensional embeddings that is relevant for sentiment can be condensed into a single dimension, hundreds of dimensions seem to be needed if we use word2vec to collect sentiment information. If we run word2vec with a small dimensionality, only a subset of available sentiment information is ``harvested'' from the corpus.

\subsection{Size of Training Resource}
Next, we analyze what size of training resource is required
\seclabel{lexicon_size_analysis} to learn a good
transformation $Q$. Labeled resources covering many words
may not be available or suffer from lack of
quality. We use the settings of lines 6 (sentiment) and 7
(concreteness) in \tabref{lexica}. \figref{lexicon_size}
shows that a small training resource of $300$ entries is
sufficient for  high performance. This suggests that
\densifier{} can create a high quality output lexicon for a
new language by hand-labeling only $300$ words; and that a
small, high-quality resource may be preferable to a large
lower-quality resource (semi-automatic or out of domain). 

To provide further evidence for this, we train \densifier{} on only the trial data of SemEval2015 task 10E. To convert  the continuous trial data to binary $-1$ / $1$ labels, we  discard all words with sentiment values between $-0.5$ and $0.5$ and round the remaining values, giving us $39$ positive and $38$ negative training words. The resulting lexicon has $\tau=$ .627 (\tabref{semeval2015_taskE}, line 8).\footnote{Here, we tune $\alpha^s$ on train (equals trial data of SemEval2015 task 10E). This seems to work  due to the different objectives for training (maximize/minimize difference) and development (correlation).} This is worse than $\tau=$ .654 (line 6) for the setup in which we used several large resources, but still better than all previous work. This indicates that \densifier{} is especially suited for languages or domains for which little training data is available.
  
\section{Related Work}\seclabel{related}

To the best of our knowledge, this paper is the first to
train an orthogonal transformation to reorder word embedding
dimensions into ultradense subspaces. However, there is much
prior work on postprocessing word
embeddings. 

\newcite{FarDodJau+15} perform postprocessing
based on a semantic lexicon with the goal of fine-tuning
word embeddings. Their transformation is not orthogonal and
therefore does not preserve distances. They show that their
approach optimizes word embeddings for a given application,
i.e., word similarity, but also that it worsens them for
other applications like detecting syntactic
relations. \newcite{FarDodJau+15}'s approach also does not
have the benefit of ultradense embeddings, in particular the
benefit of increased efficiency. 

In a tensor framework,
\newcite{rothe-schutze:2015:P15-1} transform the word
embeddings to sense (synset) embeddings. In
their work, all embeddings live in the same space whereas we
explicitly want to change the embedding space to create
ultradense embeddings with several desirable properties.

\newcite{xing-EtAl:2015:NAACL-HLT} restricted the work of
\newcite{mikolov2013exploiting} to an orthogonal
transformation to ensure that normalized embeddings stay
normalized. This transformation is learned between two
embedding spaces of different languages to exploit
similarities. They normalized word embeddings in a first
step, something that did not improve our results. 

As a
reviewer pointed out, our method is also related to
Oriented PCA \cite{Diamantaras:1996:PCN:235397}. However in
contrast to PCA a solution for Oriented PCA is not
orthogonal.

Sentiment lexicons are often created semi-automatically,
e.g., by extending manually labeled seed sets of sentiment
words or adding for each word its
syno-/antonyms. Alternatively,
words frequently cooccurring
with a seed set of manually labeled
sentiment words are added
\cite{Tur02,kiritchenko2014sentiment}. \newcite{heerschop2011sentiment}
used WordNet together with a Page\-Rank-based algorithm to
propagate the sentiment of the seed set to unknown
words. \newcite{scheible10mtsent} presented a semi-automatic
approach based on machine translation of sentiment
lexicons. The winning system of SemEval2015 10E
\cite{amir-EtAl:2015:SemEval} was based on structured
skip-gram embeddings with 600 dimensions and support vector
regression with RBF
kernels. \newcite{hamdan-bellot-bechet:2015:SemEval1}, the
second ranked team, used the average of six sentiment
lexicons as a final sentiment score, a method that cannot be
applied to low resource languages. We showed that the
lexicons created by \densifier{} achieve better performance
than other semi-automatically created lexicons.

\newcite{TanWeiYan+14} train sentiment specific embeddings by extending Collobert \& Weston's model and \newcite{tang-EtAl:2014:Coling}'s skip-gram model. The first model automatically labels tweets as positive/negative based on emoticons, a process that cannot be easily transferred to other domains like news. The second uses the Urban Dictionary to expand a small list of 350 sentiment seeds. In our work, we showed that a training resource of about the same size is sufficient without an additional dictionary. \densifier{} differs from this work in that it does not need a text corpus, but can transform existing, publicly available word embeddings. \densifier{} is independent of the embedding learning algorithm and therefore extensible to other word embedding models like GloVe \cite{PenSocMan14}, to phrase embeddings \cite{yu2015} and even to sentence embeddings \cite{NIPS2015_5950}.

\section{Conclusion}
We have introduced \densifier{}, a method that is trained to focus embeddings used for an application to an ultradense subspace that contains the information relevant for the application. In experiments on SemEval, we demonstrate two benefits of the ultradense subspace. (i) Information is preserved even if we focus on a subspace that is smaller by a factor of 100 than the original space. This means that unnecessary noisy information is removed from the embeddings and robust learning without overfitting is better supported. (ii) Since the subspace is 100 times smaller, models that use the embeddings as their input representation can be trained more efficiently and have a much smaller number of parameters. The subspace can be learned with just $80-300$ training examples, achieving state-of-the-art results on lexicon creation.

We have shown in this paper that up to three orthogonal ultradense subspaces can be created. Many training datasets can be restructured as sets of similar/dissimilar pairs. For instance, in part-of-speech tasks verb/verb pairs would be similar, verb/noun pairs dissimilar. Hence, our objective is widely applicable. In future work, we will explore the possibility of factoring all information present in an embedding into a dozen or so orthogonal subspaces. This  factorization would not change the information embeddings contain, but it would make them more compact for any given application, more meaningful and more interpretable.

The nine large \densifier{} lexicons shown in \tabref{lexica} are publicly available.\footnote{\url{www.cis.lmu.de/~sascha/Ultradense/}}

\textbf{Acknowledgments.}
We gratefully acknowledge the support of DFG:
grant 
SCHU 2246/10-1.

\bibliographystyle{naaclhlt2016}
\bibliography{feature_cat}

\end{document}